\documentclass[runningheads]{llncs}

\usepackage{eccv}

\usepackage{eccvabbrv}

\usepackage{graphicx}
\usepackage{booktabs}

\usepackage[accsupp]{axessibility}  % Improves PDF readability for those with disabilities.

\DeclareMathOperator{\EX}{\mathbb{E}}% expected value

\usepackage[dvipsnames]{xcolor}
\definecolor{BrickRed}{RGB}{170, 74, 68}
\definecolor{ForestGreen}{RGB}{34, 139, 34}
\usepackage{adjustbox}

\usepackage{multirow}

\usepackage{hyperref}

\usepackage{orcidlink}

\begin{document}

\title{Fairness Under Cover: Evaluating the Impact of Occlusions on Demographic Bias in Facial Recognition } 

% Fair Face Recognition under Occluded Faces
% O gemini sugeriu isto "Fairness & Explainability in Occluded Face Recognition: A Relevant Pixel Metric"

\titlerunning{Fairness Under Cover}

\author{Rafael M. Mamede\inst{1,2}\orcidlink{0000-0002-8313-2468} \and
Pedro C. Neto\inst{1,2}\orcidlink{0000-0003-1333-4889} \and
Ana F. Sequeira\inst{1,2}\orcidlink{0000-0002-6685-2033}}

\authorrunning{R. M. Mamede et al.}

\institute{Faculty of Engineering of the University of Porto, Porto, Portugal
\and
Institute for Systems and Computer Engineering, Technology and Science, Porto, Portugal\\
\email{\{rafael.c.maia,pedro.d.carneiro,ana.f.sequeira\}@inesctec.pt}}

\maketitle

\begin{abstract}
This study investigates the effects of occlusions on the fairness of face recognition systems, particularly focusing on demographic biases. Using the Racial Faces in the Wild (RFW) dataset and synthetically added realistic occlusions, we evaluate their effect on the performance of face recognition models trained on the BUPT-Balanced and BUPT-GlobalFace datasets. We note increases in the dispersion of FMR, FNMR, and accuracy alongside decreases in fairness according to Equilized Odds, Demographic Parity, STD of Accuracy, and Fairness Discrepancy Rate. Additionally, we utilize a pixel attribution method to understand the importance of occlusions in model predictions, proposing a new metric, Face Occlusion Impact Ratio (FOIR), that quantifies the extent to which occlusions affect model performance across different demographic groups. Our results indicate that occlusions exacerbate existing demographic biases, with models placing higher importance on occlusions in an unequal fashion, particularly affecting African individuals more severely.

\end{abstract}

\section{Introduction}
\label{sec:intro}

The ever-growing interest in machine learning-based biometric applications has raised several questions regarding the safety, trustworthiness, and potentially biased behaviour of models \cite{robinson2020face, cavazos2020accuracy}. Researchers, who initially focused on achieving great levels of recognition performance, are now investigating explainability and bias problems with great detail and interest~\cite{neto2023compressed,wang2022meta,jiang2021explainable}. This research direction has been propelled by unfortunate events regarding misclassification in criminal trials \cite{courtroom} and the inability to inform the user regarding the "why" behind the wrong prediction of the model. Even in more frequent scenarios, border control face recognition is limited by the lack of information on the model's reasoning. Additionally, the European Union (EU) General Data Protection Regulation (GDPR)~\cite{regulation2016regulation} states that users have the right to an explanation.

\begin{figure}[t]
    \centering
    \includegraphics[width=.9\linewidth,height=.5\linewidth]{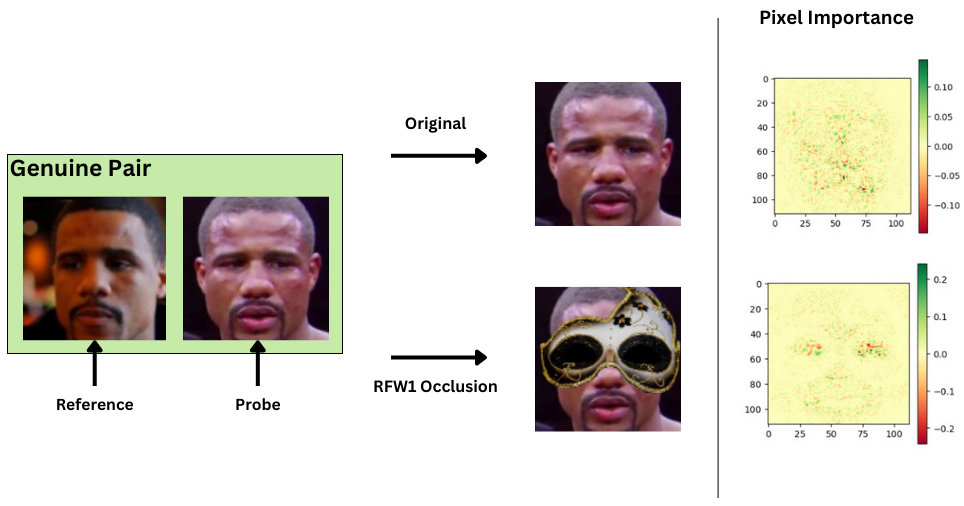}
    \caption{Genuine pair with probe image occluded, leading to incorrect classification by the Balanced34 model. Note the difference in the importance maps obtained with xSSAB \cite{huber2024efficient}: when the image is occluded most of the important pixels fall on the occluded regions. (Best viewed in color)}
    \label{fig:importance_demo}
\end{figure}

Recently, the literature has explored explainable artificial intelligence (xAI) tools to support their endeavours for efficient biases detection and potential mitigation~\cite{fu2022towards}. For instance, undesired behaviours in specific ethnic groups could be detected faster. However, the majority of explainability tools do not have knowledge of the models' inner workings~\cite{neto2024causalityinspired}. Hence, their performance as a tool can be dependent on certain demographic characteristics, as highlighted by Huber~\textit{et al.}~\cite{huber2023explainableGender}. To the best of our knowledge, these have been detected on explainability methods from the first rung of the xAI Ladder~\cite{neto2024causalityinspired}.

On Face Recognition (FR) systems, the extension of these biases and their impact is not fully understood. On the one hand, performance differences provide an indication of these biases. However, it is likely that these biases extend beyond the observed performance differences. Additionally, some sub-tasks of Face Recognition have been underdeveloped with respect to these problems. For instance, Occluded Face Recognition (OCFR), despite its resemblance to real-world scenarios where the captured image is not clear, is less studied.

Due to this lack of research on biases on OCFR, and the recent results on the potential biases mirrored by explainability tools, we have decided to align these two research topics. Specifically for OCFR, we have designed a set of experiments to measure the ethnicity-specific performance drops when a face dataset is occluded, which support our claim regarding the robustness of these systems. Additionally, we explore the usage of explainability tools to understand the predictions on these occluded datasets, as seen in figure \ref{fig:importance_demo}. Leveraging their output, we measure the rate of \textit{important} pixels that fall on top of occlusions out of all \textit{important pixels}. We further study if this information overlap is ethnicity-dependent or if it distributes uniformly across different ethnic groups. This metric provides a novel way to assess fairness in OCFR scenarios by revealing biases across sensitive groups in the system's attention on image aspects that carry no information about the identity. Hence, we can summarize our contributions as: 

\begin{itemize}
    \item Demonstrating unequal performance decrease across ethnicities in occluded scenarios;
    \item Proposing a novel way of accessing fairness when in the presence of occlusions via statistical differences of Face Occlusion Impact Ratio (FOIR) across sensitive demographics.  
    \item Demonstrating statistically significant differences in the importance of the occlusions for the predictions (FOIR) across ethnicities in False Non-Match (FNM) scenarios, that is, when a genuine pair is classified as an impostor. Notably, African examples display higher importance of the occlusions on the predictions, which aligns with the larger performance decrease seen in this ethnicity.
\end{itemize}

Additionally, we also make the occlusions used for this study available, for an easy replication of the RFW1 and RFW4 datasets\footnote{\url{https://github.com/RafaelMMamede/Fairness_OCFR}}.

\section{Related Work}

The issue of bias in FR models has garnered significant attention in recent years, with numerous studies investigating the extent and implications of these biases \cite{drozdowski2020demographic,robinson2020face}.  Grother \textit{et. al} \cite{grother2019face} reported in a comprehensive analysis of demographic differentials in FR that these algorithms are less accurate for women, older adults, and individuals of African and Asian descent compared to men and individuals of European descent. Terhörst \textit{et al.}\cite{terhorst2022comprehensive} expanded the comparison of the influence of demographics on FR, exposing the effects of other soft biometric characteristics (such as face shape and facial hair) on the verification performance of FR systems. Raji and Buolamwini \cite{raji2019actionable} have also identified systemic biases present in commercial face recognition technologies, emphasizing the need for transparency and accountability in the deployment of these systems.

Regarding ethnicity or race bias, current research highlights the different sources of biases (both from the model and the data) that can affect training \cite{wang2022meta}. The manifestation of racial bias in a face verification setting occurs when the prediction on a reference-probe pair of ethnicity \textit{A} is less likely to be wrong when compared to a reference-probe pair of a different ethnicity. In addition, the majority of works on ethnicity bias for face recognition have focused their attention on a clear evaluation setting~\cite{wang2022meta, neto2023compressed, cavazos2020accuracy}.

Recent lines of research on Fairness in FR focus not only on the detection and prevention of these biases, but also on understanding how they arise. Fu and Damer \cite{fu2022towards} reported differences in pixel attribution on key facial landmarks across demographics, hinting at different model behaviour across ethnicities. Huber \textit{et al.} \cite{huberdemographicmaps} also verified differences in the mean importance map across different demographics, with higher changes in the cases of wrongful predictions (False Matches and False Non-Matches).

One still under-explored area concerns Fairness in Occluded Face Recognition (OCFR) scenarios. OCFR deals with verification scenarios where face images are partially occluded with either realistic occlusions (such as masks, sunglasses, or scarfs \cite{neto2022beyond,neto2022ocfr}) or synthetic patches \cite{zeng2021survey}.  
Accurate identification of fairness concerns in OCFR  represents a necessity in the field of FR since, in real-world scenarios, it is common for individuals to have their faces partially covered.

Additionally, the development of better pixel attribution-based explanation tools suitable for verification scenarios, such as xSSAB\cite{huber2024efficient}, proposed by Huber \textit{et al.}, allow a better understanding of key pixels that highly contribute to the model's decision. We highly leverage these advances to develop a new metric based on the overlap of an explanation saliency map and the realistic occlusions present in the face image.

\section{Methods}
\subsection{Experimental Design}

\begin{figure}[t]
    \centering
    \includegraphics[width=.9\linewidth,height=.4\linewidth]{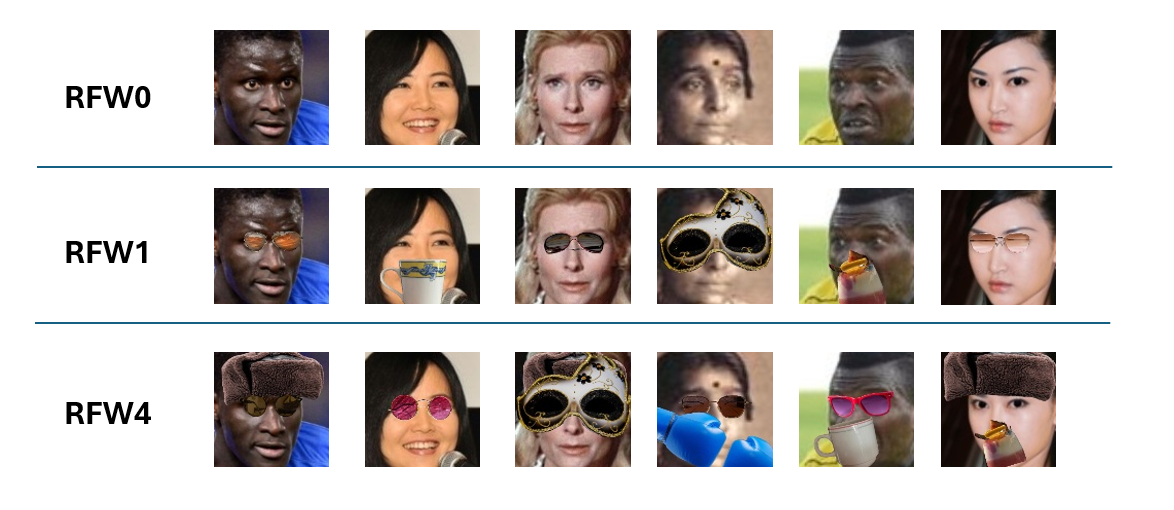}
    \caption{Examples of occlusions added with protocol 1 and protocol 4 (introduced in the 2022 Competition on Occluded Face Recognition  \cite{neto2022ocfr}) to the RFW dataset.}
    \label{fig:occlusion_demo}
\end{figure}

This study focuses on the assessment of Fairness of FR models trained and developed in clear settings, i.e. without occlusions, when faced with realistic and commonly occurring real-word occlusions. We perform our evaluation on the Racial Faces in the Wild (RFW) test dataset \cite{Wang_2019_ICCV}, containing images divided across 4 ethnicities: African, Asian, Caucasian, and Indian. We consider the proposed 6000 pairs of images for each ethnicity, 3000 of each being genuine pairs and 3000 impostor pairs. Our experimental approach can be described by the following steps:
\begin{enumerate}
\item \textbf{Generate realistic synthetic occlusions on facial images of the RFW dataset }- We start by creating the synthetic occlusions to be used throughout our benchmarks by employing two occlusion protocols, proposed in the 2022 Competition on Occluded Face Recognition \cite{neto2022ocfr} on the RFW dataset (see figure \ref{fig:occlusion_demo}). The selected protocols, 1 and 4, use affine transformations of occlusion images (e.g. carnival masks,  sunglasses, masks, etc.), warping them to fit the corresponding landmarks associated (left eye, right eye, nose, left mouth corner, and right mouth corner) identified in the RFW dataset using MTCNN \cite{zhang2016joint}.  The two protocols differ in the occlusions added, with protocol 1 adding a single occlusion (on either the upper face, lower face, eyes, or top of the head), and protocol 4 adding either a single occlusion as protocol one or two occlusions (either a combination of lower face and eye/top of head occlusions, or top of head and upper face/eye occlusions).  The choice of these protocols reflects two degrees of severity of occlusions, with protocol 4, on average, imposing more obstructions to key facial landmarks. Throughout this document, we will refer to the occluded images using protocol 1 and protocol 4 of the competition as RFW1 and RFW4, respectively. We will also refer to the unoccluded case as RFW0, aligning with the terminology for protocol 0 used in the aforementioned competition.
\item \textbf{Evaluate model performance and fairness in non-occluded and occluded scenarios} - Initially, we assess the baseline performance of the FR models (using unoccluded-unoccluded pairs of images) by measuring accuracy, False Match Rate (FMR), and False Non-Match Rate (FNMR), at a threshold optimized to maximize the difference between accuracy and standard deviation of accuracies along ethnicities (enforcing a penalization on verification thresholds with high differences of performance across ethnicities). We also evaluate a suite of Fairness metrics, described in section \ref{Metrics_section}, to determine initial disparities in the baseline case. We then utilize the synthetically occluded images to perform an assessment using unoccluded-occluded pairs of images, utilizing both protocols of occlusions. By comparing the metrics from the non-occluded and occluded scenarios, we aim to identify any changes in model performance and fairness caused by the occlusions. We also seek to understand what demographics are particularly affected by the occlusions.

\item \textbf{Utilize explanation methods to infer model behavior differences on the verification across demographics} - When dealing with verification in the unoccluded-occluded pair case, the model decision should be supported by the areas not affected by the occlusion since the occlusion is separate from the identity being observed. As such, to infer the model behavior leading to verification errors, we used a state-of-the-art efficient pixel attribution method for verification scenarios, xSSAB\cite{huber2024efficient}, to obtain the important regions for verification across our demographics. We then compute the Face Occlusion Impact Ratio (FOIR), that is, the overlap between the important pixels and the added occlusions to calculate the percentage of important pixels (IP) that fall onto occluded areas (O), $FOIR =\frac{IP\cap O}{IP}$. We look for statistical differences between the distributions of this quantity across demographics in the erroneous cases (False Matches, FM, being pairs of images not belonging to the same identify that are classified by the model as such; and False Non-Matches, FNM, which are pairs of images belonging to the same identity but classified by the model as belonging to different identities) that support the idea of distinct treatment by the model in the verification of different ethnicities. 
\end{enumerate}

\subsection{Models}

In this work, we consider models using two distinct architectures: ResNet34\cite{he2015deepresiduallearningimage} and ResNet50 \cite{he2015deepresiduallearningimage}. For each different architecture, we train the model versions using two datasets, BUPT-Balanced and BUPT-GlobalFace, proposed by Wang \textit{et al.} \cite{wang2022meta}, and intended to create a framework to study the biases of face recognition models. Each identity on the dataset has been labeled according to its skin tone into one of the following ethnicities: African, Asian, Caucasian, and Indian. BUPT-Balanced balances the number of identities that belong to each of these four categories and is composed of 1.3 million images with 28k identities, 7k identities per ethnicity. On the other hand, BUPT-Globalface contains two million images from 38k identities, and the ethnicity distribution of the identities follows the same distribution seen in the world's population. 

For the training of all models, we used the ElasticArcFace loss function (with the hyperparameters $m=0.5$, $\sigma=64$), a training batch size of 256, weight decay set to 5e-4, optimization with momentum ($\beta = 0.9$) over 26 epochs with a learning rate of 0.1 (updated by reducing an order of magnitude on epochs 8, 14, 20, and 25). We perform data augmentation on the training set with horizontal image flips.

\subsection{Group Fairness Metrics} \label{Metrics_section}

Effectively quantifying the inequalities of model adequacy in each demographic in question requires the use of well-defined fairness metrics. In this subsection, we will cover each of the metrics used in this study alongside some theoretical foundations. 

\textbf{STD of Accuracy (STD)} - The usage of the standard deviation of the verification accuracies of each protected group has been used in the field of FR as a metric for detecting disparities in quality of outcome \cite{neto2023compressed} \cite{wang2022meta}. A fairer verification procedure minimizes the STD.  

\textbf{Skewed Error Ratio (SER)} - SER is a fairness metric introduced by Wang \textit{et al.}  \cite{wang2020mitigating} in the context of FR. Considering the set of protected groups $\{g_1,g_2,...,g_n\}$, the SER is defined as:
$$SER_{@th} = \frac{\max_{g_i}Error_{@th}(g_i)}{\min_{g_i}Error_{@th}(g_i)} = \frac{\max_{g_i}(100-Acc_{@th}(g_i))}{\min_{g_i}(100-Acc_{@th}(g_i))}$$

The SER metric takes values from $1$ to $+\infty$, with lower values corresponding to lower discrepancy between the predictive performance of the groups, hence a fairer prediction.

\textbf{Fairness Discrepancy Rate (FDR) }~- FDR, proposed by Pereira {et al.} \cite{pereira2022fairness},  is a fairness evaluation metric applied in the field of FR. This metric combines the influence of both discrepancies in the False Match Rate (FMR) and the False Non-Match Rate (FNMR) in the protected groups. For our $n$ protected groups, $G =\{g_1,g_2,...,g_n\}$, FDR is given by:

$$FDR_{@th} = 1 - (\alpha A_{@th} + (1-\alpha) B_{@th}),$$
with the auxiliary ranges $A_{@th}$ and $B_{@th}$ defined as:
$$A_{@th} = \max_{g_i,g_j \in G} (|FMR_{g_i}@th - FMR_{g_j}@th|)$$

$$B_{@th} = \max_{g_i,g_j \in G} (|FNMR_{g_i}@th - FNMR_{g_j}@th|)$$

Here, $\alpha$ is a weight factor that allows for attributing the desired relative importance to the contributions for the disparity of each error rate.  The values of FDR range from 0, most unfair, to 1, most fair. 

\textbf{Inequity Rate (IR) }~- The IR is a demographic fairness metric proposed by NISP \cite{grother2022face}, in the context of FR, as an alternative to the FDR that leverages ratios between error rates rather than differences. For our $n$ protected groups, $G =\{g_1,g_2,...,g_n\}$, IR is given by:

$$IR_{@th} =  A_{@th}^{\alpha}B_{@th}^{1-\alpha},$$
with the auxiliary ranges $A_{@th}$ and $B_{@th}$ defined as:
$$A_{@th} = \frac{max_{g_i \in G} FMR_{g_i}@th}{min_{g_i \in G} FMR_{g_i}@th}$$

$$B_{@th} = \frac{max_{g_i \in G} FNMR_{g_i}@th}{min_{g_i \in G} FNMR_{g_i}@th}$$

Here, $\alpha$ is a weight factor that allows for attributing the desired relative importance to the contributions for the disparity of each error rate.  The values of FDR range from 1, most fair, to $+\infty$, most unfair. 

\textbf{Gini Aggregation Rate for Biometric Equitability (GARBE)} - GARBE is a group fairness metric proposed for biometric applications \cite{howard2022evaluating}, based on an upper bound normalized variant of the Gini coefficient. Given the formulation of the Gini coefficient, for $n$ observations of a discrete variable $x$ :
$$G_x = \frac{n}{n-1}\frac{\sum\limits_{i=1}^n\sum\limits_{j=1}^n|x_i-x_j|}{2n^2\bar{x}},$$ the GARBE metric follows as a weighted average of the gini coefficients of the samples of FMR and FNMR across each protected group:

$$GARBE_{@th} = \alpha G_{FMR@th} + (1-\alpha) G_{FNMR@th}$$

The values of GARBE range from 0 to 1, with lower values corresponding to fairer results.

\textbf{Demographic Parity (DP)} - DP \cite{agarwal2018reductions} is a fairness property that a binary model, $f$, is said to satisfy if its selection rate is independent of membership in the sensitive groups,  $P(f(X)=1|X\in g_i) = P(f(X)=1), \forall g_i \in G$. In practice, to measure whether membership in the sensitive groups affects the positive outcome of the model, we can measure the Demographic Parity Difference as:

$$DP_{@th}=\max_{g_i,g_j \in G} |\EX(f(X)|X\in g_i)-\EX(f(X)|X\in g_j)|$$

The values of DP Difference range from 0 to 1, with lower values corresponding to fairer results.

\textbf{Equalised Odds (EO)} - EO  \cite{agarwal2018reductions} is a fairness property that a binary model is said to satisfy if it performs equally well for each sensitive group. This can be seen as a more restricted version of DP that enforces the same True Positive Rate (TPR) and False Positive Rate (FPR) across all classes, $P(f(X)=1|y=v,X\in g_i) = P(f(X)=1|y=v), \forall g_i \in G, v\in \{0,1\}$ . Similar to DP, in practice, we can use the difference between TPR and FPR across all groups to obtain a proxy quantity for how well this property is satisfied. As such, the Equalised Odds Difference is defined as:

$$EO_{@th}=\max (\Delta_{TPR}@th,\Delta_{FPR}@th),$$
with $\Delta_{TPR}$ and $\Delta_{FPR}$ defined as:

$$\Delta_{TPR}@th = \max_{g_i,g_j \in G} |TPR@th_{g_i}-TPR@th_{g_j}|$$
$$\Delta_{FPR}@th = \max_{g_i,g_j \in G} |FPR@th_{g_i}-FPR@th_{g_j}|$$

The values of EO Difference range from 0 to 1, with lower values corresponding to fairer results.

\section{Experimental Results}

\subsection{Assessing Model Performance and Fairness}

In this section, we present and discuss the experimental results obtained regarding the performance and fairness of each considered model on the unoccluded and occluded scenarios. From analysing table \ref{table_fairness}, which summarizes our experimental fairness assessment, we can identify the following:
\begin{itemize}
    \item EO and DP show a large increase on both occluded scenarios across all tested models. In general, we also see that this increase tends to be larger for the more occluded RFW0-RFW4 scenario. An increase in DP difference indicates different rates of positive predictions across our demographics; however, more concerning is the increase in EO, which suggests an increase of bias in the reliability of the predictions.
    \item FDR shows small decreases in both occluded scenarios. This decrease has contributions from an increase in the range of FMR and FNMR (see table \ref{table_dispersion} and \ref{table_accuracies_and_errors}). Although there is no clear difference in the metric across protocols of occlusion, the disparity in FMR seems to be larger in the RFW0-RFW4 scenario.
    \item  STD shows an increase in both occluded scenarios; however, we do not see a clear trend on which occlusion protocol demonstrates a higher increase, varying from model to model. African examples seem to be the most affected, as they consistently show the largest accuracy decrease across all protocols and models (see table \ref{table_accuracies_and_errors} in Annex A).
    \item SER, IR, and GARBE  show a decrease in both occluded scenarios, initially pointing to a fairness increase when occlusions are added. We explore the reason and implications of these results in section \ref{Ratio}.
\end{itemize}

Also of note, the majority of each model's wrong predictions in occluded scenarios occur by incorrectly identifying a true pair as a false match (see table \ref{table_accuracies_and_errors} in Annex A). In general, FMR is less affected than FNMR across all models by adding occlusions, although we still report an increase. In both occlusion cases, the dispersion of FMR and FNMR increases across ethnicities.

\begin{table*}[t]
\centering
\caption{ Fairness of the studied models under unoccluded (RFW0-RFW0) and occluded (RFW0-RFW1, RFW0-RFW4) scenarios. The first letter on the models indicates the training set as either BUPT-Balanced (B) or BUPT-GlobalFace (G). The following number represents the model architecture, Resnet34 (34) and Resnet50 (50).
For ease of comparison, we include the percentual increase/decrease on the metric when comparing it with the unoccluded scenario. The color scheme represents whether the variation represents a \textcolor{ForestGreen}{Fairer} or \textcolor{BrickRed}{Unfairer} decision according to the corresponding metric.(Better viewed in colour)\label{table_fairness}}
\resizebox{\textwidth}{!}{ % Resize to fit within text width
\fontsize{7}{10}\selectfont % Set font size to 8pt with 10pt line spacing

\begin{tabular}{ |p{.8cm}|p{1.8cm}||
 p{1.55cm}| p{1.3cm}|  p{1.55cm}|  p{1.55cm}|p{1.2cm}|  p{1.2cm}| p{1.3cm}|}
\hline

\multicolumn{2}{|c||}{Setting}& \multicolumn{7}{|c|}{Fairness Metrics}\\
  \hline
 Model& Dataset&
 STD&SER&EO&DP&FDR&IR&GARBE \\
 \hline\hline
\multirow{3}{1.5em}{B34}&RFW0-RFW0&
1.07&1.56&0.07&0.05&0.97&3.0&0.29\\
&RFW0-RFW1&
1.92\textcolor{BrickRed}{(+79\%)}&1.29\textcolor{ForestGreen}{(\text{-}17\%)}&0.11\textcolor{BrickRed}{(+57\%)}&0.09\textcolor{BrickRed}{(+80\%)}&0.95\textcolor{BrickRed}{(\text{-}2\%)}&1.9\textcolor{ForestGreen}{(\text{-}36\%)}&0.17\textcolor{ForestGreen}{(\text{-}41\%)}\\

&RFW0-RFW4&1.52\textcolor{BrickRed}{(+42\%)}&1.17\textcolor{ForestGreen}{(\text{-}25\%)}&0.11\textcolor{BrickRed}{(+57\%)}&0.10\textcolor{BrickRed}{(+100\%)}&0.95\textcolor{BrickRed}{(\text{-}2\%)}&1.7\textcolor{ForestGreen}{(\text{-}44\%)}&0.13\textcolor{ForestGreen}{(\text{-}55\%)}\\
\hline\hline
\multirow{3}{1.5em}{G34}&RFW0-RFW0&0.66&1.30&0.04&0.03&0.98&4.9&0.25\\
&RFW0-RFW1&
1.66\textcolor{BrickRed}{(+152\%)}&1.25\textcolor{ForestGreen}{(\text{-}4\%)}&0.11\textcolor{BrickRed}{(+175\%)}&0.09\textcolor{BrickRed}{(+200\%)}&0.95\textcolor{BrickRed}{(\text{-}3\%)}&2.4\textcolor{ForestGreen}{(\text{-}52\%)}&0.17\textcolor{ForestGreen}{(\text{-}32\%)}\\
&RFW0-RFW4&
1.83\textcolor{BrickRed}{(+177\%)}&1.23\textcolor{ForestGreen}{(\text{-}4\%)}&0.15\textcolor{BrickRed}{(+275\%)}&0.12\textcolor{BrickRed}{(+300\%)}&0.94\textcolor{BrickRed}{(\text{-}4\%)}&2.1\textcolor{ForestGreen}{(\text{-}58\%)}&0.16\textcolor{ForestGreen}{(\text{-}36\%)}\\
\hline\hline
\multirow{3}{1.5em}{B50}&RFW0-RFW0&0.99&1.59&0.06&0.05&0.97&3.3&0.31\\
&RFW0-RFW1&
1.69\textcolor{BrickRed}{(+71\%)}&1.23\textcolor{ForestGreen}{(\text{-}23\%)}&0.09\textcolor{BrickRed}{(+50\%)}&0.08\textcolor{BrickRed}{(+60\%)}&0.95\textcolor{BrickRed}{(\text{-}2\%)}&2.0\textcolor{ForestGreen}{(\text{-}39\%)}&0.19\textcolor{ForestGreen}{(\text{-}39\%)}\\
&RFW0-RFW4&
2.17\textcolor{BrickRed}{(+119\%)}&1.27\textcolor{ForestGreen}{(\text{-}20\%)}&0.12\textcolor{BrickRed}{(+100\%)}&0.10\textcolor{BrickRed}{(+100\%)}&0.95\textcolor{BrickRed}{(\text{-}2\%)}&1.8\textcolor{ForestGreen}{(\text{-}47\%)}&0.15\textcolor{ForestGreen}{(\text{-}52\%)}\\
\hline\hline
\multirow{3}{1.5em}{G50}&RFW0-RFW0&0.72&1.42&0.04&0.03&0.99&4.1&0.23\\
&RFW0-RFW1&
1.55\textcolor{BrickRed}{(+115\%)}&1.23\textcolor{ForestGreen}{(\text{-}13\%)}&0.09\textcolor{BrickRed}{(+125\%)}&0.07\textcolor{BrickRed}{(+133\%)}&0.96\textcolor{BrickRed}{(\text{-}3\%)}&2.4\textcolor{ForestGreen}{(\text{-}42\%)}&0.17\textcolor{ForestGreen}{(\text{-}26\%)}\\
&RFW0-RFW4&
1.46\textcolor{BrickRed}{(+103\%)}&1.17\textcolor{ForestGreen}{(\text{-}18\%)}&0.10\textcolor{BrickRed}{(+150\%)}&0.08\textcolor{BrickRed}{(+167\%)}&0.96\textcolor{BrickRed}{(\text{-}3\%)}&2.0\textcolor{ForestGreen}{(\text{-}50\%)}&0.15\textcolor{ForestGreen}{(\text{-}35\%)}\\
\hline
\end{tabular}
}

\end{table*}

%%%NEWTABLE

\begin{table*}[t]
\centering
\caption{Auxiliary dispersion quantities $\Delta$ and $MAD$ alongside the percentual increase/decrease on the metric when compared with the unoccluded scenario. The color scheme represents whether the variation represents a \textcolor{ForestGreen}{Fairer} or \textcolor{BrickRed}{Unfairer} decision according to the corresponding metric.(Better viewed in colour)\label{table_dispersion} }
\resizebox{\textwidth}{!}{ % Resize to fit within text width
\fontsize{8}{10}\selectfont % Set font size to 8pt with 10pt line spacing
\begin{tabular}{ |p{.8cm}|p{2cm}||
p{1.7cm}|  p{1.9cm}||
p{1.7cm}|  p{1.9cm}||
p{1.7cm}|}
\hline
\multicolumn{2}{|c||}{Setting}& \multicolumn{2}{|c||}{FMR}& \multicolumn{2}{|c||}{FNMR}&Error\\
  \hline
 Model& Dataset& $\Delta_{FMR}$&$MAD_{FMR}$ & 
$\Delta_{FNMR}$&$MAD_{FNMR}$&
$\Delta_{Err}$\\
 \hline\hline
\multirow{3}{1.5em}{B34}&RFW0-RFW0&0.06& 0.030&
0.06 & 0.021&2.7\\
&RFW0-RFW1&0.11\textcolor{BrickRed}{(+83\%)}& 0.050\textcolor{BrickRed}{(+70\%)}&
0.08\textcolor{BrickRed}{(+33\%)}& 0.029\textcolor{BrickRed}{(+35\%)}&5.4\textcolor{BrickRed}{(+101\%)}\\
&RFW0-RFW4&0.11\textcolor{BrickRed}{(+83\%)}& 0.047\textcolor{BrickRed}{(+59\%)}&
0.08\textcolor{BrickRed}{(+33\%)}& 0.035\textcolor{BrickRed}{(+63\%)}&4.1\textcolor{BrickRed}{(+51\%)}\\
\hline\hline
\multirow{3}{1.5em}{G34}&RFW0-RFW0&0.04& 0.016&
0.03& 0.011&1.5\\
&RFW0-RFW1&0.11\textcolor{BrickRed}{(+175\%)}& 0.041\textcolor{BrickRed}{(+151\%)}&
0.07\textcolor{BrickRed}{(+133\%)}&0.034\textcolor{BrickRed}{(+215\%)}&4.5\textcolor{BrickRed}{(+202\%)}\\
&RFW0-RFW4&0.14\textcolor{BrickRed}{(+250\%)}& 0.055\textcolor{BrickRed}{(+234\%)}&
0.08\textcolor{BrickRed}{(+167\%)}& 0.040\textcolor{BrickRed}{(+232\%)}&5.0\textcolor{BrickRed}{(+234\%)}\\
\hline\hline
\multirow{3}{1.5em}{B50}&RFW0-RFW0&0.06&0.024&
0.06&0.021&2.6\\
&RFW0-RFW1&0.09\textcolor{BrickRed}{(+50\%)}&0.042\textcolor{BrickRed}{(+71\%)}&
0.09\textcolor{BrickRed}{(+50\%)}&0.035\textcolor{BrickRed}{(+67\%)}&4.4\textcolor{BrickRed}{(+72\%)}\\
&RFW0-RFW4&0.12\textcolor{BrickRed}{(+100\%)}&0.051\textcolor{BrickRed}{(+110\%)}&
0.10\textcolor{BrickRed}{(+67\%)}&0.038\textcolor{BrickRed}{(+81\%)}&5.9\textcolor{BrickRed}{(+128\%)}\\
\hline\hline
\multirow{3}{1.5em}{G50}&RFW0-RFW0&0.05&0.019&
0.01&0.005&1.8\\
&RFW0-RFW1&0.09\textcolor{BrickRed}{(+80\%)}&0.037\textcolor{BrickRed}{(+98\%)}&
0.06\textcolor{BrickRed}{(+500\%)}&0.028\textcolor{BrickRed}{(+451\%)}&4.0\textcolor{BrickRed}{(+126\%)}\\
&RFW0-RFW4&0.10\textcolor{BrickRed}{(+100\%)}&0.037\textcolor{BrickRed}{(+102\%)}&
0.06\textcolor{BrickRed}{(+500\%)}&0.030\textcolor{BrickRed}{(+490\%)}&3.8\textcolor{BrickRed}{(+114\%)}\\
\hline
\end{tabular}
}

\end{table*}

\subsection{Fairness Increase of Ratio-based Metrics on the Occluded Scenarios\label{Ratio}}

Initially, from interpreting the experimental results on table \ref{table_fairness}, the decrease of SER, IR, and GARBE can seem conflicting with the remaining results, as it suggests a fairness increase in the occluded scenarios. Here, we expand on these results by providing a more in-depth discussion. Firstly, we note that each of these metrics is dependent on the calculation of some ratio-based quantity related to the errors across the models. This is more easily identifiable in the SER and IR metrics, so we will start by addressing them first. 

Considering the set of protected groups $\{g_1, g_2, ..., g_n\}$, the SER metric can be rewritten, defining $\Delta_{Err@th}=\max_{g_i}(Err_{@th}(g_i))-\min_{g_i}(Err_{@th}(g_i))$, as:
$$SER = 1 + \frac{\Delta_{Err@th}}{\min_{g_i}(Err_{@th}(g_i))}$$

Since we note a high error increase across all ethnicities in the occluded verification scenarios, the SER metric is highly influenced by the fluctuation of the minimal error of the protected groups, $\min_{g_i}(Err_{@th}(g_i))$. Even though the dispersion of the error, $\Delta_{Err@th}$, increases additionally between 51\% and 234\% of the baseline unoccluded comparison (table \ref{table_dispersion}), this growth is overshadowed by the overall error growth.

Similarly, we can rewrite the auxiliary ranges used in the IR calculation as a function of the dispersion of FMR and FNMR across demographics, $\Delta_{F[N]MR@th}=\max_{g_i}(F[N]MR_{@th}(g_i))-\min_{g_i}(F[N]MR_{@th}(g_i))$ as:

$$A_{@th} = 1 + \frac{\Delta_{FMR@th}}{min_{g_i \in G} FMR_{g_i}@th}$$
$$B_{@th} = 1 + \frac{\Delta_{FNMR@th}}{min_{g_i \in G} FNMR_{g_i}@th}$$

~Here, we verify the same behaviour as for the SER metric. Even though the dispersion of FMR and FNMR, $\Delta_{FMR@th}$ and $\Delta_{FNMR@th}$, increases additionally between 42\% and 382\% of the baseline unoccluded comparison(table \ref{table_dispersion}), this increase is not enough to keep up with the overall increase of these rates.

Lastly, the Gini coefficients on GARBE are dependent on the mean absolute difference of both FMR and FNMR:

$$G_x \propto \frac{\sum\limits_{i=1}^n\sum\limits_{j=1}^n|x_i-x_j|}{n^2}, x\in\{FMR,FNMR\}$$

We show that this quantity empirically increases between 35\% and 490\% of the baseline unoccluded comparison in table \ref{table_dispersion}. Here, we observe once more that the variation increase of average FMR and FNMR affects the Gini calculations more severely, leading to a decrease of the GARBE despite an increase in dispersion.

In summary, while certain metrics like SER, IR, and GARBE indicate an increase in fairness, this is primarily driven by the overall increase in error rates. We note in all cases, the dispersion of the error or error rates increases as we add occlusions, suggesting unfairer results under occlusions. Future work should focus on refining these metrics and methodologies to ensure that they remain consistently applicable in scenarios where model errors are orders of magnitude apart, guaranteeing that improvements in fairness are genuine and not merely artifacts of increased error rates.

\subsection{Pixel Attribution in the Occluded Scenarios }

After verifying an increase in the dispersion of FNMR and FMR across ethnicities, alongside an increase in fairness indicators such as STD, EO, DP, and FDR, we seek to understand if there are significant differences in the contribution of pixels on the occlusions across ethnicities. For this we designed a novel metric, FOIR,  based on information extracted from explainability tools.
For each pair in the occluded cases, we used the xSSAB method to extract saliency maps on the occluded image. Important pixels were considered those with contributions of at least 60\% of the attribution on the most important pixel in the direction of the decision, that is we consider only negative contributions for negative predictions and positive contributions for positive predictions. These important pixels represent the areas of the image that more strongly contribute to a given model's prediction. We then calculate the overlap between important pixels and added occlusions to determine the percentage of important pixels that fall onto occluded areas (FOIR). By performing one-way analysis of variance (ANOVA) \cite{Ross2017} we seek to verify statistically significant differences in the averages of this quantity across ethnicities.

The results, on table \ref{table_overlaps}, indicate that in FNM cases, when two images of the same identity are presented, the contributions of the mask to the erroneous decision vary significantly across ethnicities. African faces show a higher degree of importance of occluded pixels in all but one tested setting, meaning that genuine pairs of this particular ethnicity are consistently more affected by occlusions. This higher overlap rate correlates with the largest increase in FNMR observed for African individuals.

Furthermore, the analysis of FM cases revealed no significant differences in the averages of FOIR across ethnicities, indicating that the model's erroneous acceptance of impostor pairs is not significantly influenced by occlusions in a manner that varies by ethnicity. This suggests that the impact of occlusions on model fairness is more pronounced in the FNM scenarios, where the model incorrectly rejects genuine pairs.

Our proposed methodology, based on the FOIR metric, can also be utilized in other comparison scenarios. In this case, we explored the wrongful predictions, however, a similar approach could explore differences of mask importance in match and non-match
predictions. Additionally, different thresholds for what constitutes an important pixel could be explored.

\begin{table}[t]
  \caption{Percentage of important pixels that fall onto occluded areas in failure cases in each occluded scenario (FOIR). P-values from ANOVA tests, indicating statistical differences in group averages (in bold for a significance level of 0.05), are provided for each scenario.\label{table_overlaps}}
\centering
\resizebox{\textwidth}{!}{ % Resize to fit within text width
\fontsize{8}{10}\selectfont % Set font size to 8pt with 10pt line spacing

\begin{tabular}
{ |p{.8cm}|p{2cm}||
p{.6cm}|  p{.6cm}|  p{.6cm}|  p{.6cm}|  p{1.4cm}||
p{.6cm}|  p{.6cm}|  p{.6cm}|  p{.6cm}| p{1.4cm}|}
\cline{3-12}
\multicolumn{2}{c}{}& \multicolumn{10}{|c|}{FOIR [\%]}
\\
  \hline
\multicolumn{2}{|c||}{Setting}& \multicolumn{5}{|c||}{FM}& \multicolumn{5}{|c|}{FNM}\\
  \hline
 Model& Dataset&
 Af&As&Ca&In&p-value& 
 Af&As&Ca&In&p-value  \\
 \hline\hline
\multirow{2}{1.5em}{B34}&RFW0-RFW1&27.8&27.8&25.1&24.1&0.17&47.6&37.0&41.9&42.4&\textbf{8.60E-14}
\\
&RFW0-RFW4&31.4&28.9&27.2&30.7&0.26&56.5&51.5&50.4&54.6&\textbf{3.11E-06}
\\
\hline\hline
\multirow{2}{1.5em}{G34}&RFW0-RFW1&29.0&24.8&21.6&24.9&0.06&46.1&37.3&41.4&40.2&\textbf{1.77E-09}
\\
&RFW0-RFW4&31.3&27.6&29.2&29.4&0.26&55.7&51.1&52.5&54.9&\textbf{3.3E-03}
\\
\hline\hline
\multirow{2}{1.5em}{B50}&RFW0-RFW1&25.2&27.1&28.6&23.2&0.26&47.7&37.6&41.5&41.6&\textbf{2.83E-13}
\\
&RFW0-RFW4&31.4&29.7&26.6&29.8&0.25&54.6&48.3&52.2&51.8&\textbf{7.53E-05}
\\
\hline\hline
\multirow{2}{1.5em}{G50}&RFW0-RFW1&24.1&25.7&18.5&21.2&0.08&41.7&34.0&38.1&39.7&\textbf{1.96E-07}
\\
&RFW0-RFW4&28.0&28.5&24.3&25.6&0.42&49.2&45.9&47.5&49.6&\textbf{0.03}
\\
\hline
\end{tabular}
}

\end{table}

\section{Conclusion}

Our investigation into the effects of occlusions on the fairness of face recognition systems has revealed significant disparities in performance across different demographic groups. The experimental results demonstrate that occlusions lead to a pronounced increase in error, with the impact being unequally distributed across ethnicities. These findings are supported by the values of the global fairness metrics  STD, EO, DP, and FDR, in both occluded scenarios. Metrics such as SER, IR, and GARBE indicate fairer decisions in occluded scenarios. However, we verified that these results are mainly due to the high increase in overall error in the occluded scenarios since the dispersion of error also increases, indicating unfairer decisions. 

Through the use of pixel attribution methods, we discovered that occlusions disproportionately affect certain ethnicities by contributing more heavily to erroneous decisions. Our analysis, utilizing the novel metric Face Occlusion Impact Ratio (FOIR), revealed that important pixels, which are critical to the model’s decision-making process, often overlap significantly with the occluded regions when dealing with genuine pairs that are misclassified by the model. This overlap was particularly pronounced in African faces compared to other ethnicities, indicating that occlusions play a larger role in the model's incorrect decisions for this group. The Face Occlusion Impact Ratio (FOIR) metric provides a vital tool for quantifying these biases, offering a clear and measurable way to assess how different demographic groups are affected by occlusions. By incorporating FOIR into the evaluation process, we can gain deeper insights into the specific ways occlusions impact model performance and fairness. 

This study highlights the need for verifying fairness not only in clean scenarios but also under commonly occurring natural occlusions. Addressing this issue is crucial for developing face recognition systems that are robust and fair across diverse demographic groups. Future research should focus on enhancing training protocols, designing more robust model architectures, and implementing comprehensive fairness evaluations that account for real-world conditions, including occlusions. By ensuring that models perform fairly under occluded conditions, we can build trust in these technologies and ensure they benefit all users equitably.

\section{Acknowledgments}

This work was developed within the Component 5 - Capitalization and Business Innovation, integrated in the Resilience Dimension of the Recovery and Resilience Plan within the scope of the Recovery and Resilience Mechanism (MRR) of the European Union (EU), framed in the Next Generation EU, for the period 2021 - 2026, within project NewSpacePortugal, with reference 11. 

\bibliographystyle{splncs04}
\bibliography{main}

\appendix

\section{Accuracies, FMR and FNMR}

\begin{table*}
\centering
\caption{ Accuracy, FMR, and FNMR of the studied models under unoccluded (RFW0-RFW0) and occluded (RFW0-RFW1, RFW0-RFW4) scenarios. \label{table_accuracies_and_errors}}
\resizebox{\textwidth}{!}{ % Resize to fit within text width
\fontsize{7}{10}\selectfont % Set font size to 8pt with 10pt line spacing

\begin{tabular}{ |p{.8cm}|p{1.8cm}||
 p{.6cm}| p{.6cm}|  p{.6cm}|  p{.6cm}||
 p{.6cm}| p{.6cm}|  p{1cm}|  p{.6cm}||
  p{.6cm}| p{.6cm}|  p{.6cm}|  p{.6cm}|}
\hline

\multicolumn{2}{|c||}{Setting}& \multicolumn{4}{|c||}{Accuracy}&
\multicolumn{4}{|c||}{FMR}& 
\multicolumn{4}{|c|}{FNMR}\\
  \hline
 Model& Dataset&
Af&As&Ca&In&
Af&As&Ca&In&
Af&As&Ca&In\\
 \hline\hline
\multirow{3}{1.5em}{B34}&RFW0-RFW0&
92.5&92.7&95.2&93.6&0.08&0.03&0.02&0.06&0.07&0.12&0.08&0.06\\
&RFW0-RFW1&
75.9&78.6&81.3&78.9&0.16&0.1&0.06&0.17&0.32&0.33&0.31&0.25\\
&RFW0-RFW4&72.3&75.4&76.4&74.7&0.18&0.15&0.09&0.20&0.38&0.34&0.38&0.30\\
\hline\hline
\multirow{3}{1.5em}{G34}&RFW0-RFW0&93.5&93.4&94.6&94.9&0.04&0.04&2.3E-03&0.02&0.09&0.09&0.11&0.08\\
&RFW0-RFW1&
77.1&79.3&81.6&80.3&0.12&0.14&0.03&0.13&0.34&0.28&0.34&0.27\\
&RFW0-RFW4&
73.0&74.7&78.0&75.9&0.16&0.20&0.06&0.17&0.38&0.30&0.38&0.31\\
\hline\hline
\multirow{3}{1.5em}{B50}&RFW0-RFW0&93.5&93.1&95.6&94.6&0.07&0.02&0.01&0.04&0.06&0.12&0.07&0.07\\
&RFW0-RFW1&
76.8&79.4&81.2&80.6&0.13&0.06&0.04&0.13&0.33&0.35&0.33&0.26\\
&RFW0-RFW4&
72.2&76.6&78.1&76.2&0.19&0.13&0.09&0.21&0.37&0.34&0.35&0.27\\
\hline\hline
\multirow{3}{1.5em}{G50}&RFW0-RFW0&94.0&94.4&95.8&95.4&0.05&0.04&3.3E-03&0.02&0.07&0.07&0.08&0.07\\
&RFW0-RFW1&
78.7&80.5&82.8&82.0&0.11&0.12&0.03&0.10&0.31&0.27&0.32&0.26\\
&RFW0-RFW4&
74.2&76.3&78.0&77.5&0.12&0.14&0.04&0.11&0.40&0.34&0.40&0.34\\
\hline
\end{tabular}
}
\end{table*}
\end{document}